\documentclass{article}

\usepackage[final,sglblindworkshop]{neurips_2025}
\workshoptitle{Foundation Models for the Brain and Body}

\usepackage[utf8]{inputenc}
\usepackage[T1]{fontenc}
\usepackage{hyperref}
\usepackage{url}

\hypersetup{
  pdftitle={DELTA: Language Diffusion-based EEG-to-Text Architecture},
  pdfauthor={MinGyu Jeon, HyoBin Kim},
  pdfsubject={EEG-to-Text, Foundation Models},
  pdfkeywords={EEG-to-Text, Residual Vector Quantization, Language Diffusion Model, Discrete Tokenization, Multimodal Brain-Language Learning}
}
\usepackage{booktabs}
\usepackage{amsfonts}
\usepackage{amsmath}
\usepackage{amssymb}
\usepackage{nicefrac}
\usepackage{microtype}
\usepackage{xcolor}
\usepackage{graphicx}
\usepackage{multirow}
\usepackage{stfloats}
\usepackage{array}

\graphicspath{{images/}}

\usepackage{natbib}

\title{$\Delta$ DELTA: Language Diffusion-based EEG-to-Text Architecture}

\author{%
  MinGyu Jeon\thanks{These authors contributed equally to this work.} \\
  MODULABS \\
  \texttt{jkmcoma7@gmail.com}
  \And
  HyoBin Kim\footnotemark[1] \\
  Sungkyunkwan University \\
  \texttt{hyobinkim@gmail.com}
}

\begin{document}

\maketitle

\begin{abstract}
Electroencephalogram (EEG)-to-text remains challenging due to high-dimensional noise, subject variability, and error accumulation in autoregressive decoding. We introduce DELTA, which pairs a Residual Vector Quantization (RVQ) EEG tokenizer with a masked language diffusion model (LLaDA). RVQ discretizes continuous EEG into multi-layer tokens to reduce noise and individual differences, while LLaDA reconstructs sentences via non-sequential denoising. On ZuCo, DELTA improves semantic alignment by up to 5.37 points over autoregressive baselines, achieving BLEU-1 21.9 and ROUGE-1 F 17.2 under word-level conditions. These results enable reliable text generation from small EEG-text datasets and point toward scalable multimodal EEG-language models.
\end{abstract}


\section{Introduction}

Brain-Computer Interfaces (BCIs) offer a vital communication channel for individuals with severe neuromuscular disorders by translating brain signals into commands for external devices\cite{chaudhary2016brain}. While early BCI research relied on invasive methods \cite{metzger2023high}, the field shifted to non-invasive EEG \cite{moses2021neuroprosthesis}, initially focusing on classification tasks \cite{autthasan2024mixnet}. More recently, Large Language Models (LLMs) have driven significant progress in EEG-to-Text translation \cite{wang2022open, dewave2023, mishra2024thought2text}. 


Despite these advancements, existing EEG-to-Text models face significant challenges in terms of reliability. The teacher-forcing technique used in the evaluation process of many studies can overestimate a model's true performance \cite{areeeg2024}, and performance measured without teacher-forcing suggests that current models have not yet sufficiently overcome the inherent low signal-to-noise ratio (SNR) problem of EEG data \cite{jiang2019removal}.

The limitations of previous research are evident from two perspectives: signal processing and generative models. First, from a signal processing standpoint, continuous EEG signals are inherently unstable. Early studies opened new possibilities with an Open-Vocabulary approach that directly mapped EEG waveforms to language models \cite{wang2022open}. However, due to the high inter-subject variability and extreme noise characteristic of EEG, such direct mapping methods can yield results akin to processing random values rather than extracting meaningful information \cite{areeeg2024}. Second, from a generative model perspective, conventional Autoregressive (AR) methods are vulnerable to error accumulation. AR models like BART generate text sequentially, using the token generated in the previous step as input for the next \cite{lewis2019bart, raffel2020t5, zhang2020pegasus}. This structure is susceptible to "error accumulation," where a small initial error can cascade and severely degrade the quality of the entire output \cite{wang2022self}. This issue becomes a critical limitation, especially in environments with noisy EEG signals, where the probability of inferring an incorrect token is higher.

To address the instability in signal processing and the error accumulation in autoregressive models, this paper proposes a novel EEG-to-Text framework. The proposed model integrates an RVQ (Residual Vector Quantization)-based EEG tokenizer with a Diffusion Model-based non-autoregressive language model. First, to tackle the high noise and inter-subject variability of EEG, we transform the continuous waveforms into a robust discrete representation using RVQ. The hierarchical quantization process of RVQ sequentially encodes the signal from its core features, effectively filtering out noise and establishing a stable foundation for extracting consistent semantic information from unstable brainwaves \cite{soundstream2021, encodec2022}. Next, to fundamentally solve the error accumulation problem of AR models, we introduce the concept of restoration from diffusion models into the text generation process \cite{ho2020denoising, austin2021structured}. Instead of generating text sequentially, this approach restores the final output by progressively denoising the entire sentence structure. This inherently prevents the sequential error propagation where an error in one step directly affects the next, enabling stable sentence generation even from noisy EEG inputs \cite{llada2025}.

The key contributions of this study are as follows: Methodological Innovation: We propose the first EEG-to-Text framework that combines an RVQ tokenizer and a diffusion-based language model, addressing challenges in both signal processing and text generation. Paradigm Shift: We shift the paradigm from the conventional direct translation approach of 'EEG-to-Text' to a restoration approach, significantly improving the stability of the generation process. Performance and Robustness Validation: We demonstrate that the proposed model achieves significant performance improvements over existing SOTA models under conditions identical to real-world inference, without relying on teacher-forcing.

\section{Method}

\begin{figure*}[t]
  \centering
  \includegraphics[width=1\textwidth]{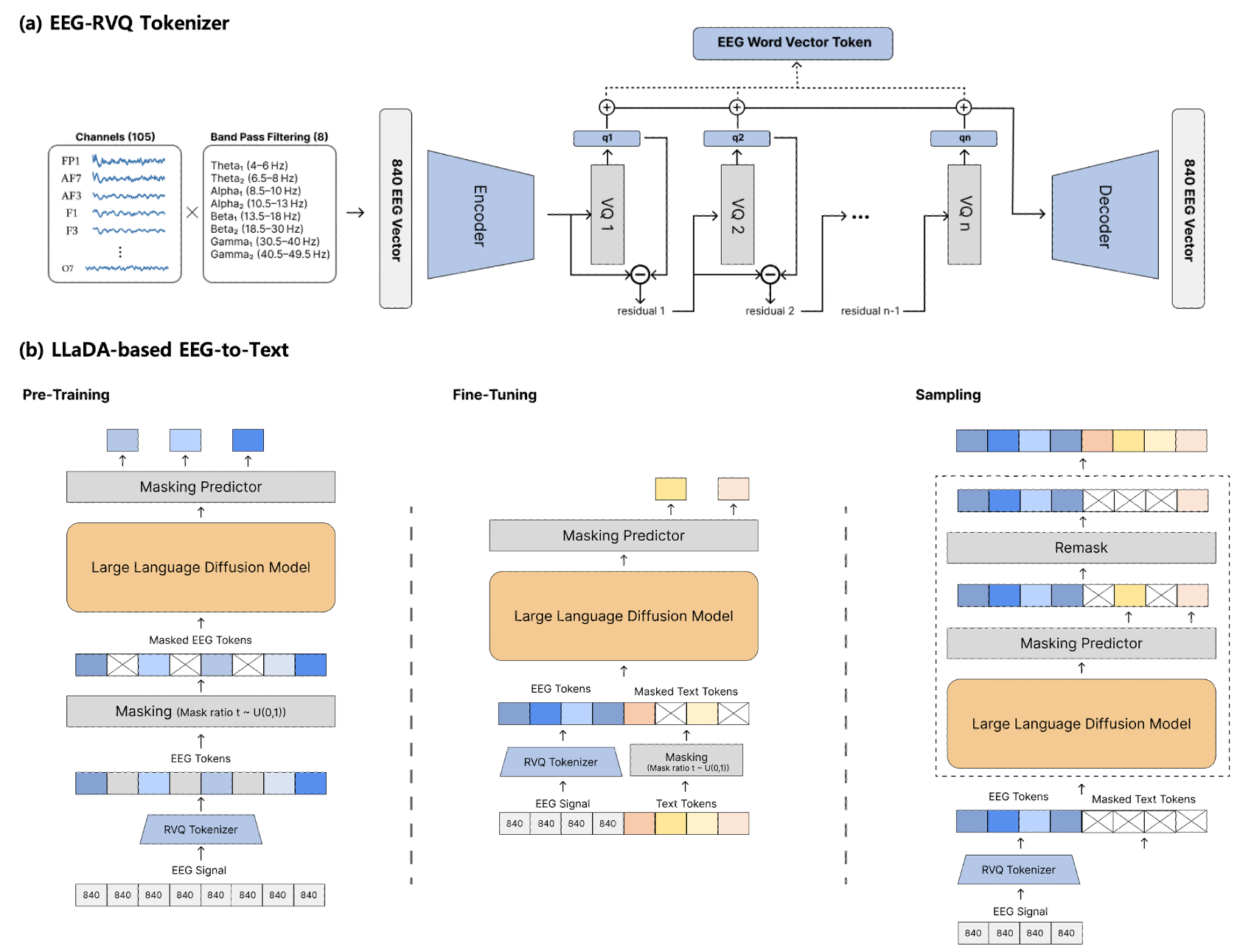}
  \caption{The DELTA framework: (a) An RVQ-based tokenizer discretizes EEG signals. (b) A diffusion model is then pre-trained on EEG tokens, fine-tuned for EEG-to-Text generation, and used for inference.}
  \label{fig1}
\end{figure*}
  

In this section, we describe the construction and training of our proposed EEG-to-Text generation framework. The framework consists of two stages: (1) RVQ-based EEG tokenizer that converts EEG signals into discrete tokens, and (2) LLaDA-based language diffusion model that generates text given the discrete tokens. The two modules are trained in stages and finally combined to generate sentences from EEG signals. Figure ~\ref{fig1} shows the overall structure schematically.

\subsection{Stage 1: RVQ-based EEG Tokenizer}

\subsubsection{EEG Feature Extraction and Quantization}
The EEG tokenizer takes continuous, multichannel EEG signals as input, extracts latent features, and quantizes them into a sequence of discrete tokens using the RVQ. In this study, we constructed an extended input tensor of 105 channels $\times$ 8 frequencies, equivalent to 840 channels, by separating 8 additional frequency bands from the original 105 channels. We then converted it into a compressed feature representation $Z$ via a one-dimensional convolutional encoder. The RVQ module then quantizes $Z$ into $M$ codebooks, generating $M$ code indices $\{q_1, q_2, \dots, q_M\}$.

RVQ obtains a high-resolution discrete representation by applying hierarchical quantization over $M$ codebooks. The final quantization vector $z_q$ is the sum of the selected code vectors from each codebook:
\begin{equation}
\begin{aligned}
z_{q} = \sum_{m=1}^{M} c_{(m, q_m)}
\end{aligned}
\end{equation}

\subsubsection{RVQ Training Objective}
The tokenizer is trained as a VQ-VAE \cite{van2017neural} by minimizing the standard objective, which combines a reconstruction loss with codebook and commitment losses to ensure a robust discrete representation:

\begin{equation}
\begin{aligned}
\mathcal{L}_{\text{VQ-VAE}} = \underbrace{\text{MSE}(x, \hat{x})}_{\text{Reconstruction Loss}} &+ \underbrace{\text{MSE}(\text{sg}(z_e), z_q)}_{\text{Codebook Loss}} \\
&+ \beta \cdot \underbrace{\text{MSE}(z_e, \text{sg}(z_q))}_{\text{Commitment Loss}}
\end{aligned}
\end{equation}

\subsection{Stage 2: LLaDA-based Text Generation}

LLaDA is a Large Language Diffusion Model, a new paradigm of language model that generates text through a stochastic masking and denoising process, unlike autoregressive models. In this study, we apply LLaDA's diffusion-based language generation technique to the EEG-to-Text problem by structuring its training and inference into distinct phases.

\subsubsection{EEG Pre-Training}
First, we perform a pre-training step using only the discrete EEG tokens obtained from the RVQ tokenizer, following the approach of Nie et al. \cite{llada2025}. This stage involves a large-scale random masking and restoration process on the EEG tokens themselves, without using any text data. By learning to restore the masked EEG tokens from their noised version, LLaDA learns the inherent distribution and structure of the EEG token space. The objective is to minimize the following loss function:
\begin{equation}
\mathcal{L}_{\text{Pre-train}} = \mathbb{E}_{t, e_0 \sim p(e)} [ \text{CE}(f_{\theta}(e_t, t), e_0) ]
\end{equation}
Here, $e_0$ is the original sequence of EEG tokens, and $e_t$ is the corrupted version of $e_0$ at a random timestep $t$. The model $f_{\theta}$ is trained to predict $e_0$ from $e_t$ and $t$. This process equips the model with diffusion-based learning capabilities tailored for EEG signals before it learns the cross-modal task of text generation.

\subsubsection{EEG-to-Text Supervised Fine-Tuning (SFT)}
After pre-training, the model is fine-tuned for the main task of generating text conditioned on EEG signals. In this stage, the model $f_{\theta}$ is given the discrete EEG tokens $\hat{Q}$ as a conditional prompt. The model's objective is to learn the probability distribution $P_\theta(Y|\hat{Q})$ to reconstruct the original text sequence $Y$ from a partially masked or noised version of it. The training process minimizes the following loss function:
\begin{equation}
\begin{aligned}
\mathcal{L}_{\text{LLaDA}} = \mathbb{E}_{t, x_0 \sim p(x)} [ \text{CE}(f_{\theta}(x_t, t, \hat{Q}), x_0) ]
\end{aligned}
\end{equation}
Here, $x_0$ represents the original, clean sequence of text tokens. $t$ is a randomly sampled timestep, and $x_t$ is the noised version of $x_0$ at that timestep. The model $f_{\theta}$ is trained to predict the original text $x_0$ given the corrupted text $x_t$, the timestep $t$, and the conditional EEG tokens $\hat{Q}$. The loss is the cross-entropy (CE) between the model's prediction and the original text.

\subsubsection{Inference}
During inference, the model generates a sentence conditioned solely on the given EEG tokens $\hat{Q}$. The process begins with a sequence of tokens that are all completely masked, representing maximum uncertainty. The model then iteratively applies a reverse diffusion (denoising) process for a fixed number of steps. In each step, the model refines its prediction for the entire sequence of text tokens, gradually filling in the masked positions to form a coherent sentence. This non-autoregressive, parallel reconstruction approach is advantageous as it minimizes the cumulative errors common in sequential prediction. An incorrectly predicted token in one step can be corrected in subsequent steps, making the generation process more robust to the noisy and variable nature of EEG signals.
\section{Experiments}

\subsection{Datasets}

In this study, we used an integrated set of EEG-text pairs from ZuCo 1.0 \cite{hollenstein2018zuco} and ZuCo 2.0 \cite{hollenstein2019zuco}. Both datasets provide raw signals recorded from 128 channels of EEG, sampled at 500 Hz, over a frequency band of approximately 0.1-100 Hz, while subjects read English sentences naturally (Normal Reading, NR) or performed a specific task (Task-Specific Reading, TSR). After subsequent denoising, 105 channels of EEG are selected as active channels, and frequency and time series characteristics are extracted for each channel, normalized to the range of 0 to 1.

Preprocessing was performed for each word read by the subject to extract EEG segments corresponding to every sentence separately. Sentence-specific EEG sequences with missing values (NaN) or extreme noise were removed. Finally, out of a total of 25,616 sentences, 20,492 (approximately 80\%) were assigned to the training set, 2,562 (approximately 10\%) to the validation set, and 2,562 (approximately 10\%) to the test set. We applied a unique sentence-based segmentation to avoid the same sentence appearing twice. By covering ZuCo data, which comprises various reading tasks (NR, TSR) and text sources, we have created an environment that enables us to validate the open lexical mapping of noisy EEG signals to natural language text in multiple ways.

\subsection{Experimental setups}

All experiments were conducted on a single NVIDIA L40S 40GB GPU using PyTorch. We used the AdamW optimizer with a batch size of 32 and employed early stopping to prevent overfitting. To manage computational resources, we fine-tuned the LLaDA-8B model in our DELTA framework using the Quantized Low-Rank Adapter (QLoRA) technique for both the EEG token pre-training and final generation stages \cite{dettmers2023qlora}. For comparison, we benchmarked our approach against autoregressive models, including BART \cite{lewis2019bart}, Pegasus \cite{zhang2020pegasus}, and T5 \cite{raffel2020t5}, which were also trained on our RVQ-tokenized EEG data. Model performance was evaluated on the test set using BLEU-N (N = 1, 2, 3, 4) \cite{papineni2002bleu}, ROUGE-1 \cite{lin2004rouge}, and WER \cite{klakow2002testing} scores.

\begin{table*}[t]
\centering
\small 
\caption{Evaluation results on sentence- and word-level features.}
\label{tab:eval}
\setlength{\tabcolsep}{4pt} 
\begin{tabular*}{\textwidth}{@{\extracolsep{\fill}} llcccccccc @{}}
\toprule
\multirow{2}{*}{Source} & \multirow{2}{*}{Method} &
\multicolumn{4}{c}{\textbf{BLEU-N (\%)}} &
\multicolumn{3}{c}{\textbf{ROUGE-1 (\%)}} &
\multirow{2}{*}{\textbf{WER (\%)}}\\
\cmidrule(lr){3-6}\cmidrule(lr){7-9}
 & & N=1 & N=2 & N=3 & N=4 & P & R & F & \\[-0.2em]
\midrule
\multirow{4}{*}{\shortstack[l]{Sentence-level features}}
 & BART              &  11.02    &  1.08   &  0.42   &  0.28   &  7.85   &  8.04   &  6.54   &  142.1 \\
 & Pegasus           &  6.20    &  0.55   &  0.24   &  0.16   &  4.90   &  5.05   &  4.03   &  147.8 \\
 & T5                &  12.50    &  1.18   &  0.48   &  0.31   &  8.6   &  8.72   &  7.15   &  141.0 \\
 & DELTA (proposed)  & 14.82 & 1.36 & 0.56 & 0.37 & 9.12 & 9.24 & 7.26 & 139.71 \\
\midrule
\multirow{4}{*}{\shortstack[l]{Word-level features}}
 & BART              & 13.69 & 2.97 & 0.82 & 0.32 & 11.98 & 13.43 & 11.87 & 108.43 \\
 & Pegasus           &  8.47 & 2.48 & 0.81 & 0.25 &  0    &  0    &  0    &  99.69 \\
 & T5                & 16.64 & 5.80 & 1.96 & 0.81 & 12.28 & 12.88 & 11.85 & 111.13 \\
 & DELTA (proposed)  & 21.93 & 6.43 & 2.01 & 0.76 & 18.88 & 18.86 & 17.24 & 110.03 \\
\bottomrule
\end{tabular*}
\end{table*}

\subsection{Qualitative Analysis}

As shown in Table~\ref{tab:eval}, our proposed DELTA model demonstrates superior performance. On \textbf{sentence-level features}, DELTA significantly outperforms autoregressive models in BLEU and ROUGE scores (e.g., 14.82 BLEU-1, 7.26 ROUGE-1 F), proving its effectiveness in restoring global context. However, its WER is high (139.71) as the sentence-level aggregation disrupts the word order information crucial for the metric.

With \textbf{word-level features}, all models improve, but DELTA's lead widens. It surpasses the next-best models, T5 and BART, by over 5 percentage points in BLEU-1 (21.93) and ROUGE-1 F (17.24), respectively. Despite being slightly behind in BLEU-4, DELTA's overall performance confirms its strong semantic restoration capabilities.

\subsection{Case Study}

A closer look at generation examples reveals two key behaviors. First, successful generations can be nearly identical to the ground truth, accurately capturing complex structures. More notably, even in semantic failures, the model preserves the original's syntactic framework. For example, one prediction incorrectly altered a sentence's subject but retained its core grammatical structure (e.g., \texttt{"(Name) (Dates) was a (Profession)."}). This indicates that our model effectively learns syntax, even when it fails to decode the correct semantic content.

\section{Conclusion}

In this study, we propose a DELTA framework that combines an RVQ-based EEG tokenizer with the LLaDA to convert continuous electroencephalogram (EEG) signals into discrete tokens with minimal information loss and generate rich contextual text through a non-sequential diffusion process to achieve superior performance on key metrics such as BLEU and ROUGE compared to existing autoregressive-based methods. These results demonstrate that language diffusion models are a promising alternative for brain-signal-based natural language generation. In the future, we plan to extend the model's generality through large-scale pre-training and integrate other brain signals, such as MEG and multimodal inputs, further to improve the accuracy and applicability of non-invasive brain-language interfaces.



\begin{ack}
This research was supported by Brian Impact Foundation, a non-profit organization dedicated to the advancement of science and technology for all.
\end{ack}

\bibliographystyle{plainnat}
\bibliography{references_v3,software}

\end{document}